\documentclass[letterpaper, 10 pt, conference]{ieeeconf}  

\IEEEoverridecommandlockouts                              

\title{\LARGE \bf
Chance Constrained Motion Planning for High-Dimensional Robots
}

\author{Siyu Dai, Shawn Schaffert, Ashkan Jasour, Andreas Hofmann, and Brian Williams
\thanks{*Computer Science and Artificial Intelligence Laboratory, Massachusetts Institute of Technology, Cambridge, MA, USA. Corresponding email {\tt\small sylviad@mit.edu}}
}

\usepackage{array}
\newcolumntype{L}[1]{>{\centering\arraybackslash}m{#1}}
\renewcommand{\tabcolsep}{3pt}
\usepackage{multirow}
\usepackage{threeparttable}
\usepackage{amssymb}
\usepackage{graphicx}
\usepackage{dblfloatfix}
\usepackage{morefloats}
\usepackage{amsmath}
\usepackage[ruled, linesnumbered, vlined]{algorithm2e}  
\newcommand{\forcond}{$i=0$ \KwTo $n$}
\SetArgSty{textnormal}
\newcommand{\llIf}[2]{{\let\par\relax\lIf{#1}{#2}}}
\newcommand{\llElse}[1]{{\let\par\relax\lElse{#1}}}

\usepackage{amsthm}   
\theoremstyle{definition}
\newtheorem{problem}{Problem}

\begin{document}

\maketitle
\thispagestyle{empty}
\pagestyle{empty}

\begin{abstract}

This paper introduces \emph{Probabilistic Chekov (p-Chekov)}, a chance-constrained motion planning system that can be applied to high degree-of-freedom (DOF) robots under motion uncertainty and imperfect state information. Given process and observation noise models, it can find feasible trajectories which satisfy a user-specified bound over the probability of collision. Leveraging our previous work in deterministic motion planning which integrated trajectory optimization into a sparse roadmap framework, p-Chekov shows superiority in its planning speed for high-dimensional tasks. P-Chekov incorporates a linear-quadratic Gaussian motion planning approach into the estimation of the robot state probability distribution, applies quadrature theories to waypoint collision risk estimation, and adapts risk allocation approaches to assign allowable probabilities of failure among waypoints. Unlike other existing risk-aware planners, p-Chekov can be applied to high-DOF robotic planning tasks without the convexification of the environment. The experiment results in this paper show that this p-Chekov system can effectively reduce collision risk and satisfy user-specified chance constraints in typical real-world planning scenarios for high-DOF robots.


\end{abstract}

\section{INTRODUCTION}


Robotic manipulation tasks in the presence of noise inevitably suffer from uncertainties and collision risks. One representative example is a manipulator mounted on an underwater vehicle, which faces not only the disturbances from currents and inner waves, but also the base movements caused by the interaction between manipulators and the vehicles on which they are mounted. Due to limited battery life, such tasks usually need to balance the risk of collisions and the costs associated with the trajectory. Another typical scenario is motion planning for human support robots, which are often surrounded by elder people and children and have to be very careful about collision avoidance. In the aforementioned applications, it is insufficient to simply rely on feedback controllers because deviations remain despite their application and there are no guarantees for task success. Therefore, for those uncertainty-sensitive planning tasks where safety and accuracy are crucial, it is important that the motion planner can reason over uncertainties during the planning phase, and react intelligently according to different planning scenarios and constraints over the probability of plan failure, i.e. \emph{chance constraints}~\cite{ono2008iterative}.

However, most existing risk-aware planners are limited to applications with low-DOF robots or simplified environments with convex obstacles, meanwhile available approaches are lacking for real-time high-dimensional planning under uncertainties. In this paper we propose \emph{Probabilistic Chekov (p-Chekov)} to fill in this gap. P-Chekov is derived from the previously introduced deterministic Chekov planner~\cite{dai2018improving} and inherits its real-time planning superiority for high-DOF robots. P-Chekov innovatively applies quadrature-based sampling for collision probability estimation, which helps obtain better estimations with a limited number of samples. It also takes advantage of the concept of \emph{risk allocation}~\cite{ono2008efficient}, where an allowed probability of failure is divided among individual constraints, to help with conflict extraction and expedite the the search for feasible solutions. Based on the estimated noise models, p-Chekov can generate feasible trajectories for high-DOF robots that satisfy a user-specified chance constraint over collision failures, which is the practical need of many real-world high-dimensional planning tasks that operate in unstructured, rapidly-changing environments.





\section{RELATED WORK}

Many existing risk-aware motion planners are based on Markov Decision Processes (MDPs)~\cite{thrun2005probabilistic, burlet2004robust, alterovitz2007stochastic} or Partially Observable MDPs (POMDPs)~\cite{kurniawati2008sarsop, van2012motion}. Despite the wide application of MDP-based approaches, many of them require discretization of the state space. Even for extensions that can handle continuous planning domains, tractability is still a common issue since they typically need partitioning or approximation of the continuous state space~\cite{ono2013probabilistic}. Another class of probabilistic planners formulates the motion planning problem into an optimization problem, for example using Disjunctive Linear Program (DLP)~\cite{blackmore2006probabilistic, blackmore2010probabilistic}. The probabilistic Sulu planner (p-Sulu) in~\cite{ono2013probabilistic} performs goal-directed planning in a continuous domain. However, since p-Sulu encodes feasible regions with linear constraint approximations, it inevitably suffers from the exponential growth of computation complexity when applied in complicated 3-dimensional environments or tasks with multiple agents.  

Risk-aware extensions of sampling-based planners are also popular in the motion planning field~\cite{luders2010chance, liu2014incremental, bry2011rapidly}. However, their applications are often limited to car-like robots in simplified environments due to their disadvantages in planning speed~\cite{dai2018improving} and collision probability estimation ability for high-DOF robots in real-world complex environments. When the robot has high dimensionality, the collision checking happens in the 3D workspace whereas the planning happens in the high-dimensional configuration space. Mapping the free workspace into the configuration space is nontrivial, which hence becomes another barrier for high-dimensional risk-aware motion planning. The methods introduced in~\cite{van2011lqg} and \cite{patil2012estimating} take advantage of elliptical level sets of Gaussian distributions and the transformation of the environment to estimate waypoint collision probabilities under Gaussian noises. Nevertheless, these methods can not be trivially extended to high-DOF applications due to the difficulty of mapping between the workspace and the configuration space.

In order to address these difficulties in high-DOF robot motion planning tasks, the p-Chekov introduced in this paper combines the ideas from the Chekov ``roadmap + TrajOpt'' planner~\cite{dai2018improving}, Linear-Quadratic Gaussian motion planning (LQG-MP)~\cite{van2011lqg}, quadrature theories~\cite{hildebrand1987introduction}, and risk allocation~\cite{ono2008efficient, ono2008iterative}. P-Chekov improves upon the isolated Chekov by extracting conflicts from the planning failures in order to guide TrajOpt~\cite{schulman2014motion} to find better solutions. It applies the LQG-MP approach to estimate the state probability distributions, but differs from LQG-MP in that it aims at generating feasible trajectories in real-time that satisfy a specified risk bound and meet a local optimality criterion, instead of choosing the minimum risk trajectory among candidate trajectories. P-Chekov relies on a quadrature-based sampling method to estimate the collision probability for individual waypoints, which mitigates the inaccuracy of random sampling and avoids the difficulty of mapping between configuration space and workspace.

\section{PROBLEM STATEMENT}  \label{ps}

P-Chekov solves the robot motion planning problems under uncertainty with a guaranteed success probability, considering temporal, spatial and dynamical constraints. Under process and observation noises, the collision rate during plan executions should not exceed a user specified chance constraint. Its real-time planning feature is key to providing robots the capability to operate effectively in unstructured, uncertain, fast-changing environments.

Let $\mathcal{X} = \mathbb{R}^{n_x}$ denote the robot \emph{state space} and $\mathcal{U} = \mathbb{R}^{n_u}$ denote the \emph{control input space}. Consider a discrete-time system with a fixed time interval $\Delta T$ corrupted by Gaussian process noises $\mathbf{m}_t$ and observation noises $\mathbf{n}_t$:

\begin{equation}
\begin{aligned}
& \mathbf{x}_t = f(\mathbf{x}_{t-1}, \mathbf{u}_{t-1}, \mathbf{m}_t), & & \mathbf{m}_t \sim \mathcal{N}(0, M_t),  \\
& \mathbf{z}_t = h(\mathbf{x}_t, \mathbf{n}_t), & & \mathbf{n}_t \sim \mathcal{N}(0, N_t), 
\end{aligned}
\end{equation}

\noindent where $\mathbf{x}_t \in \mathcal{X}$ and $\mathbf{u}_t \in \mathcal{U}$ are the robot state and control input at time step $t$. The initial state $\mathbf{x}_0 \sim \mathcal{N}(\hat{\mathbf{x}}_0, \mathbf{\Sigma}_{\mathbf{x}_0})$. 

A nominal trajectory $\Pi$ is defined as a sequence of robot states and control inputs $(\mathbf{x}^*_0, \mathbf{u}^*_0, \ldots, \mathbf{x}^*_T)$ that satisfy the deterministic dynamics model $\mathbf{x}^*_t = f(\mathbf{x}^*_{t-1}, \mathbf{u}^*_{t-1}, 0)$ for $0 < t \leq T$, where the number of time steps $T$ is a finite integer. An objective $J(\Pi)$ will be specified for each planning task. The goal state should fall in a convex goal region $\mathcal{X}^\textrm{goal}$. A valid trajectory should satisfy the temporal constraint $T \times \Delta T \leq \tau$, where $\tau$ denotes the allowed execution time, and the collision chance constraint:

\begin{equation}
 P\Bigg(\bigvee^N_{i=1} \overline{C_i}\Bigg) \leq \Delta_c,
\end{equation}

\noindent where $C_i$ constrains the trajectory from colliding into each obstacle $i = 1, \ldots, N$ and $\Delta_c \in [0, 1]$ is the user specified allowed probability of failure. Then the problem solved by p-Chekov can be expressed as:

\begin{problem}
\label{risk-aware-optimization-problem}
 \begin{equation}
\begin{aligned}
& \underset{\Pi}{\textrm{minimize}} & & J(\Pi) \\
& \textrm{subject to} & & \mathbf{x}_0 \sim \mathcal{N}(\hat{\mathbf{x}}_0, \mathbf{\Sigma}_{\mathbf{x}_0}), ~\mathbf{x}^*_T \in \mathcal{X}^\textrm{goal} \\
& & & \mathbf{x}_t = f(\mathbf{x}_{t-1}, \mathbf{u}_{t-1}, \mathbf{m}_t), & 0 < t \leq T \\
& & & \mathbf{z}_t = h(\mathbf{x}_t, \mathbf{n}_t), & 0 < t \leq T  \\
& & & \mathbf{m}_t \sim \mathcal{N}(0, M_t), & 0 < t \leq T  \\
& & & \mathbf{n}_t \sim \mathcal{N}(0, N_t), & 0 < t \leq T  \\
& & & \mathbf{x}_t \in \mathcal{X},~ \mathbf{u}_t \in \mathcal{U}, & 0 < t \leq T \\
& & & P\Bigg(\bigvee^N_{i=1} \overline{C_i}\Bigg) \leq \Delta_c \\
& & & T \times \Delta T \leq \tau \\
\end{aligned}
\end{equation}
\end{problem}

In order to guarantee the performance of this approach, we make two key assumptions. First, we assume that process and observation noises are both Gaussian distributed, and the noises on different state components are independent from each other. In real world scenarios, uncertainty due to identical and independent noises is accumulated over time, thus based on the Central Limit Theorem~\cite{hoeffding1948central}, Gaussian models should be applied. Second, we assume that both the system dynamics and observation models are either linear or can be well approximated locally by their linearizations. In real-world planning tasks, robot motions will be controlled to closely follow the planned trajectory during execution, thus the local linearizations of the non-linear dynamics are good approximations for robot motions.

\section{APPROACH: PROBABILISTIC CHEKOV}

\begin{figure}[t]
      \centering
      \includegraphics[width=\linewidth]{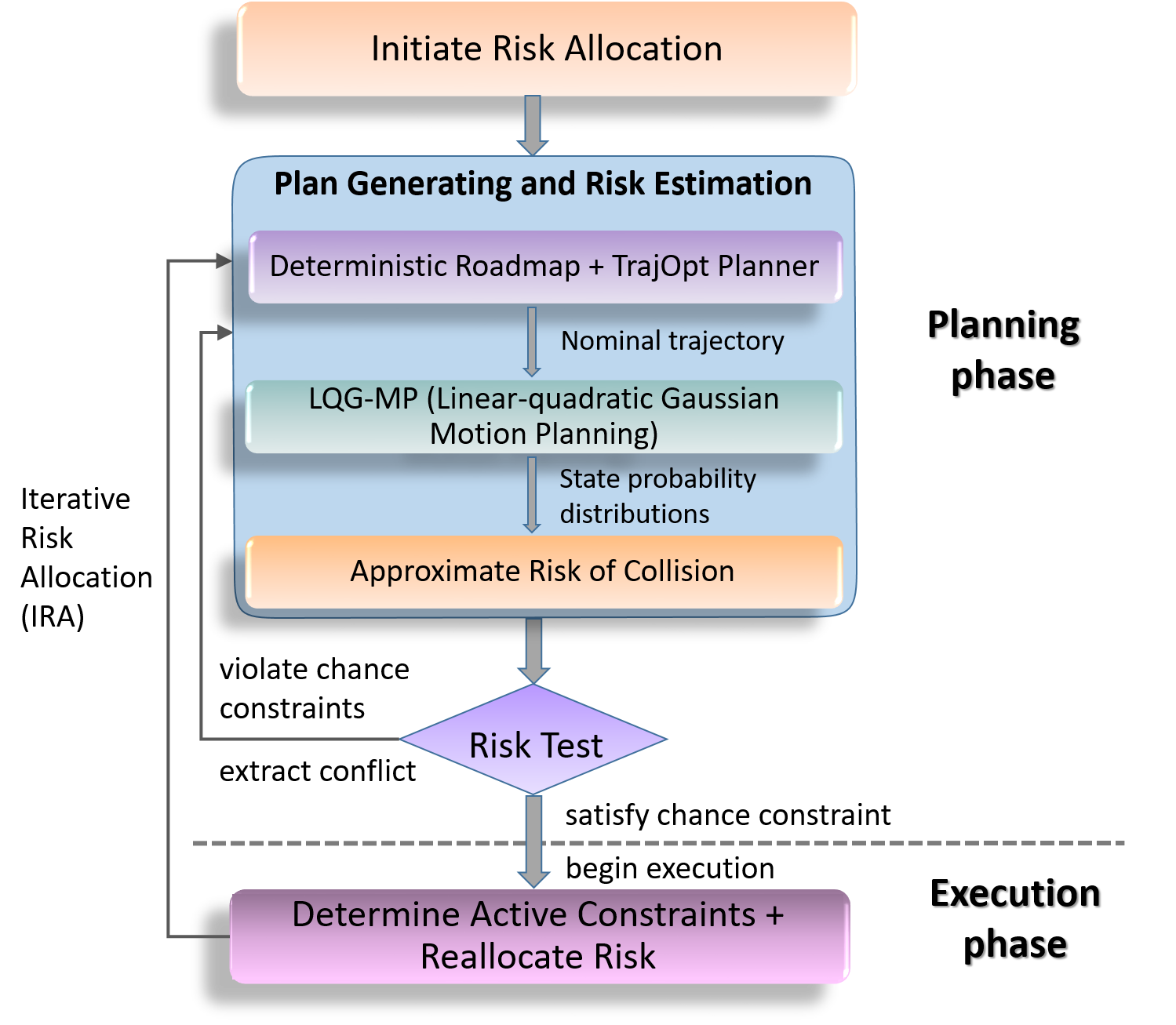}
      \caption{The p-Chekov system diagram}
      \label{fig:diagram}
   \end{figure}

Figure~\ref{fig:diagram} provides p-Chekov's system diagram, which includes a planning phase and an execution phase. The goal in the planning phase is to find an initial feasible trajectory that satisfies the given joint chance constraint $\Delta_c$. This initial trajectory is not guaranteed to be optimal, thus p-Chekov will improve it in an anytime manner during the execution phase in order to achieve better utility. However, this paper focuses on the planning phase algorithm, and detailed explanations of the execution phase is beyond the scope of this paper.

P-Chekov first uniformly distributes the chance constraint into the allowed collision risks for each waypoint along the trajectory. Using the ``roadmap + TrajOpt'' planner described in \cite{dai2018improving}, it then generates a nominal trajectory that is feasible and collision-free under deterministic dynamics. Given the model of controller and sensor noises, it estimates the \emph{a priori} probability distribution of robot states along this nominal trajectory through the LQG-MP approach~\cite{van2011lqg}. The two assumptions we made in Section~\ref{ps} ensure that the estimated states from LQG-MP are Gaussian distributed around the nominal trajectory.
The probability of collision on each waypoint can then be estimated using the quadrature-based approach in Section~\ref{quadrature_sampling}. After that, we can compare the allocated risk bound and the estimated risk of collision for each waypoint, shown as the ``risk test'' in Figure \ref{fig:diagram}. If all the risk bounds are satisfied, p-Chekov will execute this nominal trajectory. Otherwise, new constraints will be added to TrajOpt at the waypoints where the estimated risk of collision exceeds the allocated risk bound, and plan generation will be performed again. The added constraints include an increase on the \emph{penalty hit-in distance} (the minimum distance between objects where collision penalty starts to be non-zero~\cite{schulman2013finding}) and penalties on the configurations at those waypoints, so that the plan generator is guided to avoid those conflicts in future iterations. In addition, the risk allocation will also be adjusted based on the approach that will be illustrated in Section~\ref{reallocation}, so that the number of iterations it takes to find feasible trajectories could be reduced. This cycle will keep going until the solution trajectory satisfies the chance constraints or the iteration number hits its upper bound.

\subsection{Collision Probability Estimation}  \label{quadrature_sampling}


Given the state probability distribution around a nominal configuration, the collision probability can be approximated by sampling from this distribution and checking the percentage of configurations that are in collision. However, as with all Monte Carlo methods, this collision probability estimation approach would suffer from inaccuracy when the sample size is small and high computational cost when the sample size is large. Therefore, a method of intelligently finding the samples that can closely approximate the collision probability with only a small number of them is very important.

This collision probability estimation is essentially estimating the expectation of a collision function:

\begin{equation}
  c(\mathbf{x}_t) = \begin{cases} \mbox{0,} & \mbox{if } \mathbf{x}_t \mbox{ is collision free} \\ \mbox{1,} & \mbox{if } \mathbf{x}_t \mbox{ is in collision} \end{cases}
\end{equation}

\noindent along the distribution $\mathbf{x}_t \sim \mathcal{N}(\hat{\mathbf{x}}_t, \mathbf{\Sigma}_{\mathbf{x}_t})$, which is estimated using the LQG-MP approach~\cite{van2011lqg}. Since expectations can be written as integrals, non-random numerical integration methods (also called quadratures) can be applied to this problem. Denote the probability density function of $\mathbf{x}_t$ as $p(\mathbf{x}_t)$, then the expectation of $c(\mathbf{x}_t)$ can be expressed as:

\begin{equation}
 \mathbb{E}(c(\mathbf{x}_t)) = \int_{\mathbb{R}^{n_x}} c(\mathbf{x}_t)p(\mathbf{x}_t) d\mathbf{x}_t.
\end{equation}

Assume $\mathbf{x}_t$ is $d$-dimensional and let $x_t^i$ denote its $i$th component whose distributions are independent from each other, based on the independent noise assumption in Section~\ref{ps}. Since $\mathbf{x}_t$ is Gaussian distributed, we can write $x_t^i \sim \mathcal{N}(\mu_i, \sigma_i^2)$. Then, based on the conditional distribution rule of multivariate normal distribution~\cite{eaton1983multivariate}, we have:

\begin{equation}
\begin{aligned}
 & p(\mathbf{x}_t) = p(\mathbf{x}_t^{1:d}) = p(x_t^1)p(\mathbf{x}_t^{2:d}|x_t^1) = p(x_t^1)p(\mathbf{x}_t^{2:d}), \\
 & x_t^1 \sim \mathcal{N}(\mu_1, \sigma_1^2), \\
 & \mathbf{x}_t^{2:d} \sim \mathcal{N}(\mathbf{\mu}_{2:d}, \mathbf{\Sigma}_{2:d}), \\
\end{aligned}
\end{equation}

\noindent where $\mathbf{\mu}_{2:d}$ and $\mathbf{\Sigma}_{2:d}$ denote the mean and variance of $\mathbf{x}_t^{2:d}$ respectively. Since $x_t^1$ has the probability density function:

\begin{equation}
 p(x_t^1) = \frac{1}{\sigma_1 \sqrt {2\pi }} e^{- \frac{(x_t^1 - \mu_1)^2}{2 \sigma_1^2}},
\end{equation}

\noindent we can write the expectation of $c(\mathbf{x}_t)$ as:

\begin{equation}
 \mathbb{E}(c(\mathbf{x}_t)) = \int_{-\infty}^\infty p(x_t^1) \int_{\mathbb{R}^{n_x - 1}} p(\mathbf{x}_t^{2:d}) c(\mathbf{x}_t) d\mathbf{x}_t^{2:d} dx_t^1.
\end{equation}

\noindent Let $g(x_t^1) = \int_{\mathbb{R}^{n_x - 1}} p(\mathbf{x}_t^{2:d}) c(\mathbf{x}_t) d\mathbf{x}_t^{2:d}$, then:

\begin{equation}
\begin{aligned}
\label{expectation}
 \mathbb{E}(c(\mathbf{x}_t))&= \int_{-\infty}^\infty p(x_t^1)g(x_t^1)dx_t^1 \\ 
 & = \int_{-\infty}^\infty \frac{1}{\sigma_1 \sqrt {2\pi }} \exp\Big(- \frac{(x_t^1 - \mu_1)^2}{2 \sigma_1^2}\Big)g(x_t^1)dx_t^1. \\
\end{aligned}
\end{equation}

Gauss-Hermite quadrature approximates the value of integrals by calculating the weighted sum of the integrand function at a finite number of reference points, i.e.

\begin{equation}
 \int_{-\infty}^\infty e^{-y^2} h(y) dy \approx \sum_{j = 1}^n w_j h(y_j),
\end{equation}

\noindent where $n$ is the number of sampled points, $y_j$ are the roots of the Hermite polynomial $H_n(y)$ and the associated weights $w_j$ are given by~\cite{abramowitz1964handbook}:

\begin{equation}
 w_j = \frac{2^{n-1}n!\sqrt{\pi}}{n^2[H_{n-1}(y_j)]^2}.
\end{equation}

$\mathbb{E}(c(\mathbf{x}_t))$ in its form in Equation \ref{expectation} still doesn't correspond to the Hermite polynomial, therefore we conduct the following variable change:

\begin{equation}
\label{variable}
 y_1 = \frac{x_t^1 - \mu_1}{\sqrt{2}\sigma_1} \Leftrightarrow x_t^1 = \sqrt{2}\sigma_1y_1 + \mu_1.
\end{equation}

\noindent Applying Equation \ref{variable} to Equation \ref{expectation} yields:

\begin{equation}
 \mathbb{E}(c(\mathbf{x}_t)) = \int_{-\infty}^\infty \frac{1}{\sqrt{\pi}} e^{-(y_1)^2}g(\sqrt{2}\sigma_1 y_1 + \mu_1) dy_1
\end{equation}

\noindent Hence the value of $\mathbb{E}(c(\mathbf{x}_t))$ can then be approximated through Gauss-Hermite quadrature rule:

\begin{equation}
 \mathbb{E}(c(\mathbf{x}_t)) \approx \frac{1}{\sqrt{\pi}} \sum_{j = 1}^{n_1} w_{1,j} g(\sqrt{2}\sigma_1 y_{1, j} + \mu_1),
\end{equation}

\noindent where $y_{1, j}~(j = 1, \ldots, n_1)$ are the Hermite polynomial roots for integrating the $x_t^1$ component, $w_{1, j}$ are the associated weights, and $n_1$ is the number of sampled points.

If we iteratively conduct this procedure from $x_t^1$ to $x_t^d$, we will get an estimation of the collision probability through:

\begin{equation}
\label{equation:estimation}
\begin{aligned}
 \mathbb{E}(c(\mathbf{x}_t)) \approx~ & \pi^{-\frac{d}{2}} \sum_{j_1=1}^{n_1}  \sum_{j_2=1}^{n_2} \ldots \sum_{j_d=1}^{n_d} \Bigg(\prod_{i=1}^d w_{i,j_i}\Bigg) g(\sqrt{2}\sigma_1 y_{1, j_1} \\
 &  + \mu_1, \sqrt{2}\sigma_2 y_{2, j_2} + \mu_2, \ldots , \sqrt{2}\sigma_d y_{d, j_d} + \mu_d). \\
\end{aligned}
\end{equation}

\subsection{Risk Reallocation}  \label{reallocation}

Ono and Williams~\cite{ono2008efficient} introduced the concept of \emph{risk allocation}, which decomposes a joint chance constraint $\Delta$ by allocating risk bounds $\delta_i$ to individual constraints, where $\sum_1^N \delta_i = \Delta$. The problem is then separated into a risk allocation optimization stage and a control sequence optimization stage. Inspired by the concept of risk allocation and bi-stage planning, we developed a risk reallocation approach that can reduce the iterations to get feasible solutions and produce less conservative trajectories, as shown in Algorithm \ref{algo:reallocation}. 

This reallocation relies on the classification of different constraints. Denote the estimated collision risk at waypoint $i$ as $r_i$. When $r_i$ exceeds $\delta_i$, we define the chance constraint at the $i$th waypoint as a violated constraint, otherwise it is viewed as satisfied. Satisfied constraints are divided into active constraints and inactive constraints by introducing a risk tolerance parameter $\eta$. If the difference between $\delta_i$ and $r_i$ is larger than the risk tolerance, we view this chance constraint as inactive. Otherwise, the constraint is viewed as active. The key idea of this risk reallocation method is to take risk from inactive constraints and give it to those violated constraints. This is different from the Iterative Risk Allocation (IRA) algorithm introduced in~\cite{ono2008efficient} which iteratively reallocates risk from inactive constraints to active constraints. IRA requires a trajectory where all the individual chance constraints are satisfied, but it doesn't help with finding an initial satisfying trajectory. Thus it is only applicable to p-Chekov's execution phase but not the planning phase.

\begin{algorithm}
\small
\label{algo:reallocation}
 \caption{RiskReallocation}
 \DontPrintSemicolon
 
 \For{\forcond $i=1, 2, \ldots, N$}{
 \eIf{$\delta_i - r_i > \eta$}{
 $\delta_i^{new} \leftarrow \alpha\delta_i + (1 - \alpha)r_i$
 }{$\delta_i^{new} \leftarrow \delta_i$}
 }
 $\delta_{residual} = \Delta -\sum_{i=0}^N \delta_i^{new}$  \\
 $TotalViolation \leftarrow $ Sum of excessive risk from the waypoints where collision risk violates the allocated risk bound  \\
 \For{\forcond $j=1, 2, \ldots, N_{violated}$}{
 $\delta_{p_j}^{new} \leftarrow \delta_{p_j} + \delta_{residual}(r_{p_j} - \delta_{p_j})/TotalViolation$
 }
\end{algorithm}


\subsection{Probabilistic Chekov}

Algorithm~\ref{algo:pChekov} briefly summarizes the p-Chekov planning phase algorithm. The main difference between p-Chekov and other existing risk-aware motion planners relies on the usage of the ``roadmap + TrajOpt'' planner (line 1 and 6 in Algorithm~\ref{algo:pChekov}). This deterministic planner has low planning time requirements for high-DOF robots, and can straightforwardly incorporate differential constraints from robot dynamics~\cite{dai2018improving}. In each iteration, p-Chekov determines TrajOpt's penalty hit-in distances for different waypoints ($Dlist$) and the configurations to be penalized ($Plist$) based on the previous conflicts. With this ``roadmap + TrajOpt'' planner as its core, p-Chekov uses a LQG-based state estimation approach (line 7) and a quadrature-based collision probability estimation approach (line 8) in order to predict the influence of process noises and observation noises during trajectory executions. This prediction as well as the idea of risk allocation plays the role of extracting conflicts and guiding the deterministic planner to approach to a feasible solution whose execution failure rate is bounded by the chance constraint. This is the main innovation of this p-Chekov planning and execution system. In addition, the application of risk reallocation (line 14) is key to the speed of p-Chekov's convergence to a feasible solution trajectory. 


\begin{algorithm}
\small
\label{algo:pChekov}
 \caption{P-Chekov Planning Phase}
 \DontPrintSemicolon
 \SetKw{Break}{break}
 \SetKw{Return}{return}
 \KwIn{\\ $start$: \text{starting configuration of the planning query}
 \\ $goal$: \text{goal configuration of the planning query}
 \\ $\mathcal{R}$, $\mathcal{E}$: \text{robot and environment collision models}
 \\ $M_t$, $N_t$: \text{noise covariance matrices}
 \\ $\alpha$: \text{risk reallocation parameter}
 \\ $\Delta$: \text{joint chance constraint}
 \\ $\eta$: \text{risk tolerance}
 \\ $\epsilon$: \text{convergence tolerance}
 \\}
 \KwOut{\\ $\Pi$: \text{a solution trajectory}
 \\}
 
 $seed$ = RoadmapFindSolution($start, goal$) \\
 \eIf{$seed$ is not $None$}{
 Initialize penalty hit-in distances $Dlist$ with zeros \\
 Initialize risk allocation $\boldsymbol{\delta}$ with uniform allocation \\
 Initialize penalizing configurations list $Plist$ to be empty \\
 $\Pi$ = TrajOptPlanner($seed$, $Dlist$, $Plist$) \\
 $\mathcal{D} = $LQGEstimation($\Pi$, $M_t$, $N_t$) \\
 $\mathbf{r}$ = CollisionProbabilityEstimation($\Pi, \mathcal{D}, \mathcal{R}, \mathcal{E}$) \\
 $violation$ = RiskTest($\mathbf{r}, \boldsymbol{\delta}$) \\
 \While{$violation$ is $True$}{
 \ForEach{\forcond waypoint $i$ that violates risk bound}{
 Add the configuration at waypoint $i$ to $Plist$ \\
 Increase the $i$th item of $Dlist$ by $d_{step}$ \\
 }
 $\boldsymbol{\delta}$ = RiskReallocation($\mathbf{r}, \boldsymbol{\delta}, \alpha, \Delta, \eta$) \\
 $\Pi$ = TrajOptPlanner($\Pi$, $Dlist$, $Plist$) \\
 $violation$ = RiskTest($\mathbf{r}, \boldsymbol{\delta}$) \\
 }
 Chance constraint satisfied, start execution \\
 }
 {\Return Failure}
\end{algorithm}

\section{EXPERIMENTS AND RESULTS}

In order to test the performance of p-Chekov, we conducted simulation experiments on Baxter~\cite{BaxterRobot} in two tabletop environments that were previously developed in~\cite{dai2018improving}: a ``tabletop with a pole'' and a ``tabletop with a container'' environment. The test cases in the second environment are much more difficult due to the narrow spaces inside the container. In addition, unlike the ones in the first environment, these cases include difficult poses where the joints are close to their limits. 500 pairs of start and goal poses in each environment were selected to compare the performance of p-Chekov and deterministic Chekov. The test cases where the estimated collision risk of either the start or the goal has already exceeded 150\% of the chance constraint were filtered out, since those cases are very likely to be infeasible. We set the chance constraint as 10\% and run 100 executions to test the solutions for each of the 500 test cases.

\subsection{Experiment Models}

We assume all the joints are fully actuated, and linearize the manipulator dynamics around its nominal trajectory. The control inputs are the accelerations at each time step. We assume the joint dynamics are independent from each other, corrupted by process noise $\mathbf{m}_{t, j} \sim \mathcal{N}(0, M_{t, j})$, where $j = 1, 2, \ldots, 7$ denotes the DOF index. We define:

\begin{equation}
\label{deviation}
 \begin{aligned}
  & \bar{\mathbf{x}}_t = \mathbf{x}_t - \mathbf{x}^*_t, \\
  & \bar{\mathbf{u}}_t = \mathbf{u}_t - \mathbf{u}^*_t \\
 \end{aligned}
\end{equation}

\noindent as the deviations of states and inputs from the nominal trajectory. Use $\mathbf{x}_{t, j}$ to denote the state variable of the $j$th joint, then the dynamics for each joint can be linearized as:
%

\begin{equation}
 \bar{\mathbf{x}}_{t, j} = \begin{bmatrix}
                 1 & \Delta T \\
                 0 & 1 \\
                \end{bmatrix}
 \bar{\mathbf{x}}_{t-1, j} + \begin{bmatrix}
                    \Delta T^2/2  \\ \Delta T \\
                    \end{bmatrix}
 \bar{\mathbf{u}}_{t-1, j} + \mathbf{m}_{t, j}.
\end{equation}

In manipulation tasks, the relative spatial relationship between the end-effector and the object to be grabbed is important for task success. The transformation matrix between workspace objects and the end-effector can be expressed as:

\begin{equation}
 T_{obj}^{ee} = T_{obj}^{cam} \cdot T_{cam}^{ee},
\end{equation}

\noindent where $T_{obj}^{cam}$ is the transformation from the workspace object to the camera frame, and $T_{cam}^{ee}$ is the transformation from the camera frame to the end-effector. Therefore, the noises for observing $T_{obj}^{ee}$ can be transformed into observation noises for $T_{cam}^{ee}$ through the transformation matrix $T_{obj}^{cam}$. Then $T_{cam}^{ee}$ can be transformed into $T_{ee}^{cam}$ through matrix inversion. Therefore, we can approximate the observation noises by corrupted observations of the end-effector pose through the camera.


The observations of the end-effector can be expressed in joint space through the nonlinear relationship:

\begin{equation}
\begin{aligned}
& \mathbf{z}_t = h(\mathbf{x}_t, \mathbf{n}_t), & & \mathbf{n}_t \sim \mathcal{N}(0, N_t),
\end{aligned}
\end{equation}

\noindent where $h(\mathbf{x}_t, 0)$ is the forward kinematics, $\mathbf{n}_t$ is the observation noise, and $N_t$ is the covariance matrix of the observation noise. The linearization of this observation model around the nominal trajectory point $\mathbf{x}^*_t$ can be expressed as:

\begin{equation}  
\label{Jacobian}
\begin{aligned}
& \mathbf{z}_t - h(\mathbf{x}^*_t, 0) = J_t(\mathbf{x}_t - \mathbf{x}^*_t) + W_t\mathbf{n}_t,
\end{aligned}
\end{equation}

\noindent where

\begin{equation}
 \begin{aligned}
  & J_t = \frac{\partial h}{\partial \mathbf{x}}(\mathbf{x}^*_t, 0). \\
 \end{aligned}
\end{equation}

\noindent Since $h(\mathbf{x}_t, 0)$ is the forward kinematics, $J_t$ is the end-effector Jacobian matrix at the nominal state $\mathbf{x}^*_t$. In this way, the system observation matrix becomes the Jacobian matrix, which is usually easy to obtain during computation.

If we define deviations from the nominal observations as:

\begin{equation}
 \begin{aligned}
  & \bar{\mathbf{z}}_t = \mathbf{z}_t - h(\mathbf{x}^*_t, 0), \\
 \end{aligned}
\end{equation}

\noindent then the linearized system observation model shown in Equation \ref{Jacobian} can be rewritten as:

\begin{equation}
\label{observation}
\begin{aligned}
& \bar{\mathbf{z}}_t = J_t\bar{\mathbf{x}}_t + W_t\mathbf{n}_t, & & \mathbf{n}_t \sim \mathcal{N}(0, N_t),
\end{aligned}
\end{equation}

\noindent which is the end-effector observation model we need.


\subsection{Experiment Results}

%
%

The analysis of experiment results focuses on p-Chekov's comparison with deterministic Chekov and its chance constraint satisfaction performance. Within 100 independent executions for a particular solution trajectory which has a 10\% probability of collision, the number of failure cases follows a binomial distribution with the number of independent experiments $n = 100$ and the probability of occurrence in each experiment $p = 0.1$. The cumulative probability distribution function of binomial distributions can be expressed as:

\begin{equation}
 F(k; n, p) = \textrm{Pr}(X \leq k) = \displaystyle\sum_{i}^{k} {n\choose i} p^i(1-p)^{n-i}.
\end{equation}

\noindent Then we can calculate that having less than or equal to 10 failures out of 100 executions has a probability of 56\%, whereas the probability of having within 15 failures is 94\%. Therefore, we decide to define chance constraint satisfied test cases as the ones where the collision rate is lower than or equal to 150\% of the chance constraint. In this way, we have much more confidence to say that a solution violates the chance constraint when there are more than the corresponding number of executions end up in collision.

Since theoretically p-Chekov only has probabilistic guarantees for waypoints in a trajectory, we distinguish between continuous-time and discrete-time chance constraint satisfaction performances. If the 100 noisy executions of a test cases shows that the average continuous-time (or waypoint) collision rate is within 150\% of the collision chance constraint, then we say this case satisfies the continuous-time (or discrete-time) chance constraint. Only the continuous-time satisfaction is the true criterion for success, but we use discrete-time performance to show the impact of edge collisions, i.e. the collisions in between waypoints.

\begin{figure}[t]
      \centering
      \includegraphics[width=\linewidth]{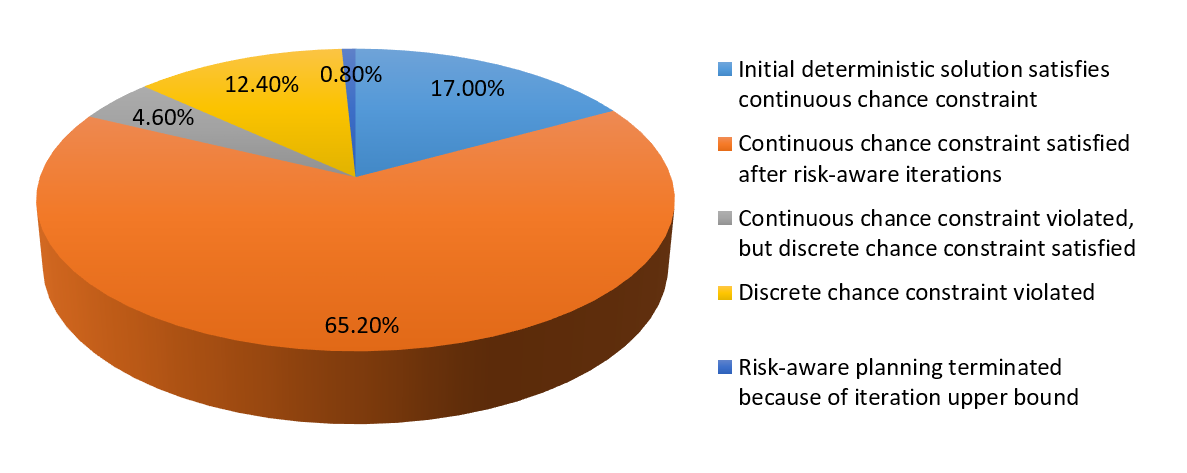}
      \caption{The statistical breakdown for Tabletop with a container environment}
      \label{fig:feasible_2}
   \end{figure}

Figure~\ref{fig:feasible_2} shows a statistical breakdown for the results in the ``tabletop with a container'' environment with a 10\% chance constraint and a 50 iteration limit. Here the continuous-time chance constraint satisfaction rate is altogether 82.20\%, including 17.00\% where the initial deterministic solution has already satisfied the chance constraint. 4.60\% of the cases fail for edge collisions, which is an inevitable outcome of the discretization of trajectories. Hence the balance between edge collision and computation complexity is crucial when deciding the number of waypoints. 

Table~\ref{table:feasible_ee_obs} provides detailed results of p-Chekov's performance in both environments. The first six rows compare the performance of deterministic Chekov (roadmap + TrajOpt) and p-Chekov's planning phase algorithm, while the remaining fourteen focus on p-Chekov's chance constraint satisfaction performance. On average, p-Chekov takes longer to plan, and returns paths with longer execution trajectories.
This is expected because it adjusts the deterministic solution iteratively in order to satisfy the chance constraint and can often push the solution trajectory further away from the locally optimal solution. The eighth and eleventh rows show that the failure cases where p-Chekov struggles to find feasible solutions cost more time, whereas the success cases have a much shorter average planning time.

\begin{table}
\caption{P-Chekov Test Results with 10\% Chance Constraint}
\label{table:feasible_ee_obs}
\centering
\begin{threeparttable}
\begin{tabular}{L{1.65cm}|L{1.15cm}|L{2.97cm}|L{0.9cm}|L{0.9cm}}
\hline
 \multicolumn{3}{c|}{Environment}  &  Tabletop with a Pole  & Tabletop with a Container   \\ 

\hline
\multirow{2}{1.65cm}{\centering Planning Time (s)} 
& \multicolumn{2}{c|}{deterministic Chekov} & 1.10  & 1.27 \\ 
& \multicolumn{2}{c|}{p-Chekov} & 19.34 & 31.17 \\ 

\hline
\multirow{2}{1.65cm}{\centering Overall Collision Rate\tnote{1}} 
& \multicolumn{2}{c|}{deterministic Chekov} & 27.51\%  & 41.04\%  \\ 
& \multicolumn{2}{c|}{p-Chekov} & 11.39\% & 16.46\%   \\ 

\hline
\multirow{2}{1.65cm}{\centering Average Path Length (rad)\tnote{2}} 
& \multicolumn{2}{c|}{deterministic Chekov} & 0.51  & 0.60 \\ 
& \multicolumn{2}{c|}{p-Chekov} & 0.68  & 0.84  \\

\hline
\multirow{12}{1.65cm}{\centering P-Chekov Performance} 
& \multicolumn{2}{p{4.22cm}|}{continuous chance constraint satisfaction rate\tnote{3}} & 87.60\% & 82.20\%  \\ \cline{2-5}
& \multirow{3}{1.15cm}{continuous satisfied cases\tnote{4}} & average planning time (s) & 15.40 & 22.95 \\ 
&  & average collision rate & 0.08\%  & 0.11\%  \\ 
&  & average risk reduction\tnote{6} & 0.25 & 0.33 \\ \cline{2-5}
& \multirow{3}{1.15cm}{continuous violated cases} & average planning time (s) & 46.22  & 67.12  \\
&  & average collision rate & 88.50\%  & 88.02\%  \\ 
&  & average risk reduction & -0.44  & -0.13  \\ \cline{2-5}

& \multicolumn{2}{p{4.22cm}|}{discrete chance constraint satisfaction rate\tnote{5}} & 94.40\% & 86.80\%  \\ \cline{2-5}
& \multirow{3}{1.15cm}{discrete satisfied cases} & average planning time (s) & 18.76 & 25.15 \\ 
&  & average collision rate & 0.13\%  & 0.10\%  \\ 
&  & average risk reduction & 0.19 & 0.28 \\ \cline{2-5}
& \multirow{3}{1.15cm}{discrete violated cases} & average planning time (s) & 28.08  & 68.78  \\
&  & average collision rate & 73.39\%  & 86.59\%  \\ 
&  & average risk reduction & -0.39  & -0.23  \\ 
\hline
\end{tabular}
\begin{tablenotes}
 \item[1] Average collision rate over 100 noisy executions for all 500 test cases.
 \item[2] Average length of execution trajectories.
 \item[3] Percentage of test cases where the average continuous-time collision rate over 100 noisy executions satisfies the chance constraint.
 \item[4] P-Chekov performance over the test cases where chance constraint is satisfied by continuous-time collision rate (viewed as success cases).
 \item[5] Percentage of test cases where the average waypoint collision rate over 100 noisy executions satisfies the chance constraint.
 \item[6] The difference between the average collision rate of p-Chekov solutions and that of deterministic Chekov solutions.
 \end{tablenotes}
 \end{threeparttable}
\end{table}

The comparison of the overall collision rate, the average over all the 100 executions for 500 cases, shows the superiority of p-Chekov solutions. The overall collision rate in the ``tabletop with a pole'' environment is within 150\% of the chance constraint, and in the ``tabletop with a container'' environment it exceeds by only a small amount. Row nine and row twelve show that this excess is mainly from the failure cases. The collision rate of success cases are very low, which means p-Chekov's planning phase produces conservative solutions, and that's why an anytime planning improving approach is in demand in the execution phase.

As shown in the seventh row, the continuous-time chance constraint satisfaction rates are above 80\% in both environments. For discrete-time chance constraint the percentages are both above 85\%. In addition, the tenth row tells us that in the satisfied cases p-Chekov successfully reduces the average collision rate by over 0.25. The increased collision risk in failure cases is probably caused by p-Chekov's struggle to find safe solutions that satisfy the chance constraint in these difficult cases, where the trajectories are pushed into some obstacles while p-Chekov is trying to get around others.

\section{DISCUSSION}

%
This paper introduced p-Chekov, a chance constrained motion planning and execution system that can be applied to high-DOF robotic systems under motion uncertainty and imperfect state information. Through the ``roadmap + TrajOpt'' framework as well as the quadrature-based risk estimation and the risk reallocation approaches, it overcame existing risk-aware planners' limitation in real-time motion planning tasks with high-DOF robots in 3-dimensional non-convex environments. Simulation tests showed that p-Chekov had high chance constraint satisfaction rate and showed a much lower collision rate compared with deterministic solutions.

However, solutions from p-Chekov's planning phase can be overly conservative due to the suboptimal risk allocations as well as the limited number of quadrature nodes. Although not included in this paper, preliminary experimental results show that applying an IRA-based anytime plan improving algorithm to the execution phase can effectively improve the solution utility. The completion of this IRA-based execution phase algorithm is an important track of our future work. Other future work directions include improving the collision probability estimation algorithm, developing more intelligent constraints that guide p-Chekov to avoid conflicts, and incorporating risk information into the roadmap nodes as a heuristic to search for low risk seed trajectories.

\addtolength{\textheight}{-11cm}   



%
%
%
%
%
%
%

\bibliographystyle{IEEEtran}
\bibliography{ICRA_2019}

\begin{thebibliography}{10}
\providecommand{\url}[1]{#1}
\csname url@rmstyle\endcsname
\providecommand{\newblock}{\relax}
\providecommand{\bibinfo}[2]{#2}
\providecommand\BIBentrySTDinterwordspacing{\spaceskip=0pt\relax}
\providecommand\BIBentryALTinterwordstretchfactor{4}
\providecommand\BIBentryALTinterwordspacing{\spaceskip=\fontdimen2\font plus
\BIBentryALTinterwordstretchfactor\fontdimen3\font minus
  \fontdimen4\font\relax}
\providecommand\BIBforeignlanguage[2]{{%
\expandafter\ifx\csname l@#1\endcsname\relax
\typeout{** WARNING: IEEEtran.bst: No hyphenation pattern has been}%
\typeout{** loaded for the language `#1'. Using the pattern for}%
\typeout{** the default language instead.}%
\else
\language=\csname l@#1\endcsname
\fi
#2}}

\bibitem{ono2008iterative}
M.~Ono and B.~C. Williams, ``Iterative risk allocation: A new approach to
  robust model predictive control with a joint chance constraint,'' in
  \emph{Decision and Control, 2008. CDC 2008. 47th IEEE Conference on}.\hskip
  1em plus 0.5em minus 0.4em\relax IEEE, 2008, pp. 3427--3432.

\bibitem{dai2018improving}
S.~Dai, M.~Orton, S.~Schaffert, A.~Hofmann, and B.~C. Williams, ``Improving
  trajectory optimization using a roadmap framework.'' in \emph{Proceedings of
  the 2018 IEEE/RSJ International Conference on Intelligent Robots and Systems
  (IROS)}, 2018, (accepted).

\bibitem{ono2008efficient}
M.~Ono and B.~Williams, ``An efficient motion planning algorithm for stochastic
  dynamic systems with constraints on probability of failure.'' 2008.

\bibitem{thrun2005probabilistic}
S.~Thrun, W.~Burgard, and D.~Fox, \emph{Probabilistic robotics}.\hskip 1em plus
  0.5em minus 0.4em\relax MIT press, 2005.

\bibitem{burlet2004robust}
J.~Burlet, O.~Aycard, and T.~Fraichard, ``Robust motion planning using markov
  decision processes and quadtree decomposition,'' in \emph{Robotics and
  Automation, 2004. Proceedings. ICRA'04. 2004 IEEE International Conference
  on}, vol.~3.\hskip 1em plus 0.5em minus 0.4em\relax IEEE, 2004, pp.
  2820--2825.

\bibitem{alterovitz2007stochastic}
R.~Alterovitz, T.~Sim{\'e}on, and K.~Y. Goldberg, ``The stochastic motion
  roadmap: A sampling framework for planning with markov motion uncertainty.''
  in \emph{Robotics: Science and systems}, vol.~3, 2007, pp. 233--241.

\bibitem{kurniawati2008sarsop}
H.~Kurniawati, D.~Hsu, and W.~S. Lee, ``Sarsop: Efficient point-based pomdp
  planning by approximating optimally reachable belief spaces.'' in
  \emph{Robotics: Science and systems}, vol. 2008.\hskip 1em plus 0.5em minus
  0.4em\relax Zurich, Switzerland., 2008.

\bibitem{van2012motion}
J.~Van Den~Berg, S.~Patil, and R.~Alterovitz, ``Motion planning under
  uncertainty using iterative local optimization in belief space,'' \emph{The
  International Journal of Robotics Research}, vol.~31, no.~11, pp. 1263--1278,
  2012.

\bibitem{ono2013probabilistic}
M.~Ono, B.~C. Williams, and L.~Blackmore, ``Probabilistic planning for
  continuous dynamic systems under bounded risk,'' \emph{Journal of Artificial
  Intelligence Research}, vol.~46, pp. 511--577, 2013.

\bibitem{blackmore2006probabilistic}
L.~Blackmore, H.~Li, and B.~Williams, ``A probabilistic approach to optimal
  robust path planning with obstacles,'' in \emph{American Control Conference,
  2006}.\hskip 1em plus 0.5em minus 0.4em\relax IEEE, 2006, pp. 7--pp.

\bibitem{blackmore2010probabilistic}
L.~Blackmore, M.~Ono, A.~Bektassov, and B.~C. Williams, ``A probabilistic
  particle-control approximation of chance-constrained stochastic predictive
  control,'' \emph{IEEE transactions on Robotics}, vol.~26, no.~3, pp.
  502--517, 2010.

\bibitem{luders2010chance}
B.~Luders, M.~Kothari, and J.~How, ``Chance constrained rrt for probabilistic
  robustness to environmental uncertainty,'' in \emph{AIAA guidance,
  navigation, and control conference}, p. 8160.

\bibitem{liu2014incremental}
W.~Liu and M.~H. Ang, ``Incremental sampling-based algorithm for risk-aware
  planning under motion uncertainty,'' in \emph{Robotics and Automation (ICRA),
  2014 IEEE International Conference on}.\hskip 1em plus 0.5em minus
  0.4em\relax IEEE, 2014, pp. 2051--2058.

\bibitem{bry2011rapidly}
A.~Bry and N.~Roy, ``Rapidly-exploring random belief trees for motion planning
  under uncertainty,'' in \emph{Robotics and Automation (ICRA), 2011 IEEE
  International Conference on}.\hskip 1em plus 0.5em minus 0.4em\relax IEEE,
  2011, pp. 723--730.

\bibitem{van2011lqg}
J.~Van Den~Berg, P.~Abbeel, and K.~Goldberg, ``Lqg-mp: Optimized path planning
  for robots with motion uncertainty and imperfect state information,''
  \emph{The International Journal of Robotics Research}, vol.~30, no.~7, pp.
  895--913, 2011.

\bibitem{patil2012estimating}
S.~Patil, J.~Van Den~Berg, and R.~Alterovitz, ``Estimating probability of
  collision for safe motion planning under gaussian motion and sensing
  uncertainty,'' in \emph{Robotics and Automation (ICRA), 2012 IEEE
  International Conference on}.\hskip 1em plus 0.5em minus 0.4em\relax IEEE,
  2012, pp. 3238--3244.

\bibitem{hildebrand1987introduction}
F.~B. Hildebrand, \emph{Introduction to numerical analysis}.\hskip 1em plus
  0.5em minus 0.4em\relax Courier Corporation, 1987.

\bibitem{schulman2014motion}
J.~Schulman, Y.~Duan, J.~Ho, A.~Lee, I.~Awwal, H.~Bradlow, J.~Pan, S.~Patil,
  K.~Goldberg, and P.~Abbeel, ``Motion planning with sequential convex
  optimization and convex collision checking,'' \emph{The International Journal
  of Robotics Research}, vol.~33, no.~9, pp. 1251--1270, 2014.

\bibitem{hoeffding1948central}
W.~Hoeffding, H.~Robbins, \emph{et~al.}, ``The central limit theorem for
  dependent random variables,'' \emph{Duke Mathematical Journal}, vol.~15,
  no.~3, pp. 773--780, 1948.

\bibitem{schulman2013finding}
J.~Schulman, J.~Ho, A.~X. Lee, I.~Awwal, H.~Bradlow, and P.~Abbeel, ``Finding
  locally optimal, collision-free trajectories with sequential convex
  optimization.'' in \emph{Robotics: science and systems}, vol.~9, no.~1.\hskip
  1em plus 0.5em minus 0.4em\relax Citeseer, 2013, pp. 1--10.

\bibitem{eaton1983multivariate}
M.~L. Eaton, ``Multivariate statistics: a vector space approach.'' \emph{JOHN
  WILEY \& SONS, INC., 605 THIRD AVE., NEW YORK, NY 10158, USA, 1983, 512},
  1983.

\bibitem{abramowitz1964handbook}
M.~Abramowitz and I.~A. Stegun, \emph{Handbook of mathematical functions: with
  formulas, graphs, and mathematical tables}.\hskip 1em plus 0.5em minus
  0.4em\relax Courier Corporation, 1964, vol.~55.

\bibitem{BaxterRobot}
RethinkRobotics, ``Baxter,'' http://www.rethinkrobotics.com/baxter/.

\end{thebibliography}

\end{document}